\documentclass[a4paper]{article}

\usepackage[utf8]{inputenc}
\usepackage[pdftex]{graphicx}
\usepackage{epstopdf}
\usepackage{authblk}
\usepackage{algorithm2e}
\usepackage{amsmath}
\usepackage{amsfonts}
\usepackage{amssymb}
\usepackage{listings}
\usepackage{tikz}
\usepackage{scalerel}
\usepackage[strings]{underscore}
\usepackage[hyphens]{url}
\usepackage[misc]{ifsym}

\usetikzlibrary{bayesnet}

\title{From abstract items to latent spaces to observed data and back: Compositional Variational Auto-Encoder}
\date{~}
\author[1]{Victor Berger}
\author[1]{Michele Sebag}
\affil[1]{TAU, CNRS $-$ INRIA $-$ Univ. Paris-Saclay, France}

\DeclareMathOperator*{\E}{\mathbb{E}}

\DeclareUnicodeCharacter{2212}{-}

\begin{document}

\maketitle

\setcounter{footnote}{0}

\begin{abstract}
    Conditional Generative Models are now acknowledged an essential tool in Machine Learning. This paper focuses on their control. While many approaches aim at disentangling the data through the coordinate-wise control of their latent representations, another direction is explored in this paper. The proposed CompVAE handles data with a natural multi-ensemblist structure ({\em i.e.} that can naturally be decomposed into elements). Derived from Bayesian variational principles, CompVAE learns a latent representation leveraging both observational and symbolic information. A first contribution of the approach is that this latent representation supports a compositional generative model, amenable to multi-ensemblist operations (addition or subtraction of elements in the composition). This compositional ability is enabled by the invariance and generality of the whole framework w.r.t. respectively, the order and number of the elements. The second contribution of the paper is a proof of concept on synthetic 1D and 2D problems, demonstrating the efficiency of the proposed  approach. 
\end{abstract}


\section{Introduction}

Representation learning is at the core of machine learning, and even more so since the inception of deep learning \cite{bengio_representation_2013}. As shown by e.g., \cite{chen_infogan:_2016,higgins_beta-vae:_2016}, the latent representations built to handle high-dimensional data can effectively support desirable functionalities. One such functionality is the  ability to directly control the observed data through the so-called representation disentanglement, especially in the context of computer vision and image processing  \cite{prasad_survey_2012,liu_deep_2018} (more in section \ref{sec:soa}).

This paper extends the notion of representation disentanglement from a latent coordinate-wise  perspective to a semi-structured setting. Specifically, we tackle the ensemblist setting where a datapoint can naturally be interpreted as the combination of multiple parts. 
The contribution of the paper is a {generative model}  built on the Variational Auto-Encoder principles \cite{kingma_auto-encoding_2013,rezende_stochastic_2014}, {\em controlling the data generation from a description of its parts} and supporting ensemblist operations such as the addition or removal of any number of parts.

The applicative motivation for the presented approach, referred to 
as {\em Compositional Variational AutoEncoder} (CompVAE),
is the following. In the domain of Energy Management, a key issue is to simulate the consumption behavior of an ensemble of consumers, where each household consumption is viewed as an independent random variable following a distribution law defined from the household characteristics, and the household consumptions are possibly correlated through external factors such as the weather, or a football match on TV (attracting members of some but not all households). Our long term goal is to infer a simulator, taking as input the household profiles and their amounts: it should be able to simulate their overall energy consumption and account for their correlations. The data-driven inference of such a programmable simulator is a quite desirable alternative to the current approaches, based on Monte-Carlo processes and requiring either to explicitly model the correlations of the elementary random variables, or to proceed by rejection.

Formally, given the description of datapoints and their parts, the goal of CompVAE is to learn the distribution laws of the parts (here, the households) and to sample the overall distribution defined from a varying number of parts (the set of households), while accounting for the fact that the parts are not independent, and the sought overall distribution depends on shared external factors: {\em the whole is not the sum of its parts}. 

The paper is organized as follows. Section \ref{sec:soa} briefly reviews related work in the domain of generative models and latent space construction, replacing our contribution in context. Section \ref{sec:model} gives an overview of CompVAE, extending the VAE framework to multi-ensemblist settings. Section \ref{sec:experiments} presents the experimental setting retained to establish a proof of concept of the approach on two synthetic problems, and section \ref{sec:resu} reports on the results. Finally section \ref{sec:discussion} discusses some perspectives for further work and  applications to larger problems.

\section{Related Work}

\label{sec:soa}
Generative models, including VAEs \cite{kingma_auto-encoding_2013,rezende_stochastic_2014} and GANs \cite{goodfellow_generative_2014}, rely on an embedding from the so-called latent space $Z$ onto the dataspace $X$. In the following, data space and observed space are used interchangeably. It has long been observed that continuous or discrete operations in the latent space could be used to produce interesting patterns in the data space. For instance, the linear interpolation between two latent points $z$ and $z'$ can be used to generate a morphing between their images \cite{radford_unsupervised_2015}, or the flip of a boolean coordinate of $z$ can be used to add or remove an elementary pattern (the presence of glasses or mustache) in the associated image \cite{dumoulin_adversarially_2016}.

The general question then is to control the flow of information from the latent to the observed space and to make it actionable. Several approaches, either based on information theory or on supervised learning have been proposed to do so.
Losses inspired from the Information Bottleneck \cite{tishby_information_2000,shwartz-ziv_opening_2017,achille_emergence_2017} and enforcing the independence of the latent and the observed variables, conditionally to the relevant content of information, have been proposed: enforcing the decorrelation of the latent coordinates in $\beta$-VAE \cite{higgins_beta-vae:_2016}; aligning the covariances of latent and observed data in \cite{kumar_variational_2017}; decomposing the latent information into pure content and pure noise in InfoGAN \cite{chen_infogan:_2016}.
Independently, explicit losses have been used to yield conditional distributions in conditional GANs \cite{mirza_conditional_2014}, or to enforce the scope of a latent coordinate in \cite{kulkarni_deep_2015,tran_disentangled_2017}, (e.g. modeling the light orientation or the camera angle). 

The structure of the observed space can be mimicked in the latent space, to afford
expressive yet trainable model spaces; in Ladder-VAE \cite{sonderby_ladder_2016}, a sequence of dependent latent variables are encoded and reversely decoded to produce complex observed objects. Auxiliary losses are added in \cite{maaloe_auxiliary_2016} in the spirit of semi-supervised learning.  In \cite{kingma_semi-supervised_2014}, the overall generative model involves a classifier, trained both in a supervised way with labeled examples and in an unsupervised way in conjunction with a generative model.

An important case study is that of sequential structures:  \cite{chung_recurrent_2015}
considers fixed-length sequences and loosely mimics an HMM process, where latent variable $z_i$ controls the observed variable $x_i$ and the next latent $z_{i+1}$. In \cite{hsu_unsupervised_2017}, a linear relation among latent variables $z_i$ and $z_{i+1}$ is enforced; in \cite{co-reyes_self-consistent_2018}, a recurrent neural net is
used to produce the latent variable encoding the current situation. In a more general context, \cite{webb_faithful_2017} provides a generic method for designing an appropriate inference network that can be associated with a given Bayesian network representing a generative model to train.

The injection of explicit information at the latent level can be used to support "information surgery" via loss-driven information parsimony. 
For instance in the domain of 
signal generation \cite{chorowski_unsupervised_2019}, the neutrality of the latent representation w.r.t. the locutor identity is enforced by directly providing the identity at the latent level: as $z$ does not need to encode the locutor information, the information parsimony pressure ensures $z$ independence wrt the locutor.
Likewise, {\em fair} generative processes can be enforced by directly providing the sensitive information at the latent level \cite{zemel_learning_2013}. In
\cite{louizos_variational_2015}, an adversarial mechanism based on Maximum Mean Discrepancy \cite{gretton_kernel_2012} is used to enforce the neutrality of the latent. In \cite{moyer_invariant_2018}, the minimization of the mutual information is used in lieu of an adversary.

\paragraph{Discussion.} All above approaches (with the except of sequential settings \cite{chung_recurrent_2015,hsu_unsupervised_2017}, see below) handle the generation of a datapoint as a whole naturally involving diverse facets; but not composed of inter-related parts. Our goal is instead to tackle the proper parts-and-whole structure of a datapoint, where the {\em whole is not necessarily the simple sum of its parts} and the parts of the whole are interdependent. 
In sequential settings \cite{chung_recurrent_2015,hsu_unsupervised_2017}, the dependency of the elements in the sequence are handled through parametric restrictions (respectively considering fixed sequence-size or linear temporal dependency) to enforce 
the proper match of the observed and latent spaces. 
A key contribution of the proposed CompVAE is to tackle the parts-to-whole
structure with no such restrictions, and specifically accommodating a varying number of parts $-$ possibly different between the training and the generation phases.

\section{Overview of CompVAE}

\label{sec:model}

This section describes the CompVAE model, building upon the VAE principles \cite{kingma_auto-encoding_2013} with the following difference: 
CompVAE aims at building a {\em programmable generative model} $p_\theta$, taking as input the ensemble of the parts of a whole observed datapoint. 
A key question concerns the latent structure most appropriate to reflect the ensemblist nature of the observed data. The proposed structure (section \ref{sec:archi}) involves a latent variable associated to each part of the whole. The aggregation of the part is achieved through an order-invariant operation, and the interactions among the parts are modeled at an upper layer of the latent representation. 

In encoding mode, the structure is trained from the pairs formed by a whole, and an abstract description of its parts; the latent variables are extracted along an iterative non-recurrent process, oblivious of the order and number of the parts (section \ref{sec:training}) and defining the encoder model $q_\phi$. \\
In generative mode, the generative model is supplied with a set of parts, and accordingly generates a consistent whole, where variational effects operate jointly at the part and at the whole levels.

\paragraph{Notations.} A datapoint $x$ is associated with an ensemble of parts noted $\{\ell_i\}$. Each $\ell_i$ belongs to a finite set of categories $\Lambda$. Elements and parts are used interchangeably in the following. In our illustrating example, a consumption curve $x$ involves a number of households; the $i$-th household is associated with its consumer profile $\ell_i$, with $\ell_i$ ranging in a finite set of profiles. Each profile in $\Lambda$ thus occurs 0, 1 or several times. The generative model relies on a learned distribution $p_\theta(x | \{ \ell_i \})$, that is decomposed into latent variables: a latent variable named $w_i$ associated to each part $\ell_i$, and a common latent variable $z$.

\def\wtil{{\mbox{$\widetilde{w}$}}}
\def\elltil{{\mbox{$\tilde{l}$}}}

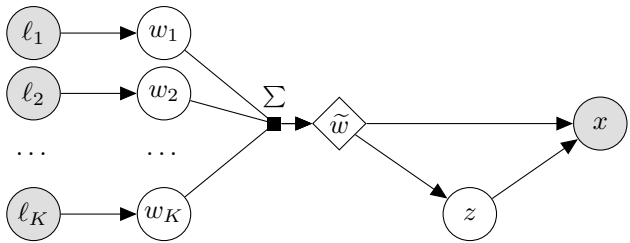
\begin{figure}[htbp]
	\centering
	\begin{tikzpicture}
    \node[obs](x){$x$};
    \node[latent, left=of x, yshift=-1.2cm](z){$z$};
    \node[det, left=of z, yshift=1.2cm](w){$\widetilde{w}$};
    \factor[left=of w]{sum}{$\sum$}{}{};
    \node[latent, left=of sum, yshift=1.2cm](w1){$w_1$};
    \node[obs, left=of w1](l1){$\ell_1$};
    \node[latent, left=of sum, yshift=0.4cm](w2){$w_2$};
    \node[obs, left=of w2](l2){$\ell_2$};
    \node[latent, left=of sum, yshift=-1.2cm](wk){$w_K$};
    \node[obs, left=of wk](lk){$\ell_K$};
    
    \edge{w, z}{x};
    \edge{w}{z};
    \factoredge{w1}{sum}{w};
    \factoredge{w2}{sum}{w};
    \factoredge{wk}{sum}{w};
    \edge{l1}{w1};
    \edge{l2}{w2};
    \edge{lk}{wk};
    \path (l2) -- node[auto=false]{\ldots} (lk);
    \path (w2) -- node[auto=false]{\ldots} (wk);
\end{tikzpicture}
	\caption{Bayesian network representation of the CompVAE generative model.}
	\label{fig:decoder-net}
\end{figure}

\subsection{CompVAE: Bayesian architecture}
\label{sec:archi}\label{sec:model-desc}\label{sec:property}\label{sec:conjec}
The architecture proposed for CompVAE is depicted as a graphical model on Fig. 
\ref{fig:decoder-net}. As said, the $i$-th part belongs to category $\ell_i$ and is associated with a latent variable $w_i$ (different parts with same category are associated with different latent variables). The ensemble of the $w_i$s is aggregated into an intermediate latent variable \wtil. 
A key requirement is for \wtil\ to be {\em invariant} w.r.t. the order of elements in $x$. In the following \wtil\ is set to the sum of the $w_i$, $\wtil = \sum_i w_i$. Considering other order-invariant aggregations is left for further work. \\
The intermediate latent variable \wtil\ is used to condition the $z$ latent variable; both \wtil\ and $z$ condition the observed datapoint $x$. This scheme corresponds to the following factorization of the generative model $p_\theta$: 
\begin{equation}
\label{eq:decoder-factorization}
p_\theta(x, z, \{w_i\} | \{\ell_i\}) = p_\theta(x|z, \wtil) p_\theta(z | \wtil) \prod_i p_\theta(w_i | \ell_i)
\end{equation}
In summary, the distribution of $x$ is conditioned on the ensemble $\{\ell_i\}$ as follows: The $i$-th part of $x$ is associated with a latent variable $w_i$ modeling the generic distribution of the underlying category $\ell_i$ together with its specifics. Variable \wtil\ is deterministically computed to model the aggregation of the $w_i$, and finally $z$ models the specifics of the aggregation.

Notably, each $w_i$ is linked to a single $\ell_i$ element, while $z$ is global, being conditioned from the global auxiliary \wtil. The rationale for introducing $z$ is to enable a more complex though still learnable distribution at the $x$ level $-$ compared with the alternative of conditioning $x$ only on \wtil. It is conjectured that an information-effective distribution would store in $w_i$ (respectively in $z$) the {\em local information} related to the $i$-th part (resp. the {\em global information} describing the interdependencies between all parts, e.g. the fact that the households face the same weather, vacation schedules, and so on). Along this line, it is conjectured that the extra information stored in $z$ is limited compared to that stored in the $w_i$s; we shall return to this point in section \ref{sec:goalexperiment}.

The property of invariance of the distribution w.r.t. the order of the $\ell_i$ is satisfied by design. A second desirable property regards the robustness of the distribution w.r.t. the varying number of parts in $x$. More precisely, two requirements are defined. The former one, referred to as {\em size-flexibility property}, is that the number $K$ of parts of an $x$ is neither constant, nor bounded {\em a priori}. The latter one, referred to as {\em size-generality property} is the generative model $p_\theta$ to accommodate larger numbers of parts than those seen in the training set.

\subsection{Posterior inference and loss}\label{sec:training}
Letting $p_D(x | \{\ell_i\})$ denote the empirical data distribution, the learning criterion to optimize is the data likelihood according to the sought generative model $p_\theta$: $\E_{p_D} \log p_\theta(x | \{\ell_i\})$. \\
The (intractable) posterior inference of the model is approximated using the Evidence Lower Bound (ELBO) \cite{jordan_introduction_1999},
following the Variational AutoEncoder approach \cite{kingma_auto-encoding_2013,rezende_stochastic_2014}. Accordingly, we proceed by optimizing a lower bound of the log-likelihood of the data given $p_\theta$, which is equivalent to minimizing an upper bound of the Kullback-Leibler divergence between the two distributions :

\begin{equation}
D_{KL}(p_D \| p_\theta) \leq H(p_D) + \E_{x \sim p_D} \mathcal{L}_{ELBO}(x)
\end{equation}

The learning criterion is, with $q_\phi(z, \{w_i\} | x, \{\ell_i\})$ the inference distribution:

\begin{equation}
\label{eq:elbo1}
\begin{split}
\mathcal{L}_{ELBO}(x) &= \E_{z, \{w_i\} \sim q_\phi} \log \frac{q_\phi(z, \{w_i\} | x, \{\ell_i\})}{p_\theta(z | \wtil)\prod_i p_\theta(w_i|\ell_i)} \\
& - \E_{z, \{w_i\} \sim q_\phi} \log p_\theta(x | z, \wtil)
\end{split}
\end{equation}

The inference distribution is further factorized as $q_\phi(\{w_i\}|z,x, \{l_i\})q_\phi(z|x)$, yielding the final training loss:

\begin{equation}
\label{eq:elbo2}
\begin{split}
\mathcal{L}_{ELBO}(x) &= \E_{z, \{w_i\} \sim q_\phi} \log \frac{q_\phi(\{w_i\} | x, z, \{\ell_i\})}{\prod_i p_\theta(w_i|\ell_i)} \\
& + \E_{z, \{w_i\} \sim q_\phi} \log \frac{q_\phi(z | x)}{p_\theta(z | \wtil)} \\
& - \E_{z, \{w_i\} \sim q_\phi} \log p_\theta(x | z, \wtil)
\end{split}
\end{equation}

The training of the generative and encoder model distributions is described in Alg.  \ref{alg:train-model}.

\begin{algorithm}
	$\theta, \phi \leftarrow \text{Random initialization}$\;
	\While{Not converged}{
		$x, \{\ell_i\} \leftarrow \text{Sample minibatch}$\;
		$z \leftarrow \text{Sample from\ } q_\phi(z | x)$\;
		$\{w_i\} \leftarrow \text{Sample from\ } q_\phi(\{w_i\} | x, z, \{\ell_i\})$\;
		$\mathcal{L}_w \leftarrow D_{KL}(q_\phi(\{w_i\} | x, z, \{\ell_i\}) \| \Pi_i p_\theta(w_i|\ell_i))$\;
		$\mathcal{L}_z \leftarrow \log \frac{q_\phi(z | x)}{p_\theta(z | \tilde{w})}$\;
		$\mathcal{L}_x \leftarrow - \log p_\theta(x | z, \tilde{w})$\;
		$\mathcal{L}_{ELBO} \leftarrow \mathcal{L}_w + \mathcal{L}_z + \mathcal{L}_x$\;
		$\theta \leftarrow \text{Update}(\theta, \nabla_\theta \mathcal{L}_{ELBO})$\;
		$\phi \leftarrow \text{Update}(\phi, \nabla_\phi \mathcal{L}_{ELBO})$\;
	}
	\caption{CompVAE Training Procedure.}
	\label{alg:train-model}
\end{algorithm}

\subsection{Discussion}
In CompVAE, the sought distributions are structured as a Bayesian graph (see $p_\theta$ in Fig. \ref{fig:decoder-net}), where each node is associated with a neural network and a probability distribution family, like for VAEs. This neural network takes as input the parent variables in the Bayesian graph, and outputs the parameters of a distribution in the chosen family, e.g., the mean and variance of a Gaussian distribution. The reparametrization trick \cite{kingma_auto-encoding_2013} is used to back-propagate gradients through the sampling.

A concern regards the training of latent variables when considering Gaussian distributions. A potential source of instability in CompVAE comes from the fact that the Kullback-Leibler divergence between $q_\phi$ and $p_\theta$ (Eq. (\ref{eq:elbo2})) becomes very large when the variance of some variables in $p_\theta$ becomes very small\footnote{Single-latent variable VAEs do not face such problems as the prior distribution $p_\theta(z)$ is fixed, it is not learned.}. To limit this risk, some care is exercised in parameterizing the variances of the normal distributions in $p_\theta$ to making them lower-bounded.

\subsubsection{Modeling of $q_\phi(\{w_i\}|x,z,\{\ell_i\})$.}
\label{sec:q-model}

The latent distributions $p_\theta(z|\wtil)$, $p_\theta(w_i|\ell_i)$ and $q_\phi(z|x)$ are modeled using diagonal normal distributions as usual. Regarding the model $q_\phi(\{w_i\} | z, x, \{\ell_i\})$, in order to be able to faithfully reflect the generative model $p_\theta$,  it is necessary to introduce the correlation between the $w_i$s in $q_\phi(\{w_i\} | z, x, \{\ell_i\})$ \cite{webb_faithful_2017}.

As the aggregation of the $w_i$ is handled by considering their sum, it is natural to handle their correlations through a multivariate normal distribution over the $w_i$. The  proposed parametrization of such a multivariate is as follows. Firstly, correlations operate in a coordinate-wise fashion, that is,  $w_{i,j}$ and $w_{i',j'}$ are only correlated if $j = j'$. The parametrization
(detailed in appendix \ref{app:multivariate-parameter})
of the $w_i$s ensures that: i) the variance of the sum of the $w_{i,j}$ can be controlled and made arbitrarily small in order to ensure an accurate VAE reconstruction; ii) the Kullback-Leibler divergence between $q_\phi(\{w_i\}|x, z, \{\ell_i\})$ and $\prod_i p_\theta(w_i|\ell_i)$ can be defined in closed form. 

The learning of $q_\phi(\{w_i\}|x, z, \{\ell_i\})$ is done using a fully-connected graph neural network \cite{scarselli_graph_2009} leveraging graph interactions akin message-passing \cite{gilmer_neural_2017}. The graph has one node for each element $\ell_i$, and every node is connected to all other nodes. The state of the $i$-th node is initialized to $(pre_\phi(x), z, e_\phi(\ell_i) + \epsilon_i)$, where $pre_\phi(x)$ is some learned function of $x$ noted, $e_\phi(\ell_i)$ is a learned embedding of $\ell_i$, and $\epsilon_i$ is a random noise used to ensure the differentiation of the $w_i$s. The state of each node of the graph at the $k$-th layer is then defined by its $k-1$-th layer state and the aggregation of the state of all other nodes:

\begin{equation}
\begin{cases}
h_i^{(0)} = (pre_\phi(x), z, e_\phi(\ell_i) + \epsilon_i) \\
h_i^{(k)} = f_\phi^{(k)}\left(h_i^{(k-1)}, \sum_{j \neq i} g_\phi^{(k)}(h_j^{(k-1)})\right)
\end{cases}
\end{equation}

where $f_\phi^{(k)}$ and $g_\phi^{(k)}$ are learned neural networks: $g_\phi^{(k)}$ is meant to embed the current state of each node for an aggregate summation, and $f_\phi^{(k)}$ is meant to "fine-tune" the $i$-th node conditionally to all other nodes, such that they altogether account for \wtil.

\section{Experimental Setting}

\label{sec:experiments}
This section presents the goals of experiments and describes the experimental setting used to empirically validate CompVAE. 

\subsection{Goals of experiments}
\label{sec:goalexperiment}

As said, CompVAE is meant to achieve a programmable generative model. From a set of latent values $w_i$, either derived from $p_\theta(w_i|\ell_i)$ in a generative context, or recovered from some data $x$, it should be able to generate values $\hat{x}$ matching any chosen subset of the $w_i$. This property is what we name the "ensemblist disentanglement" capacity, and the first goal of these experiments is to investigate whether CompVAE does have this capacity.

A second goal of these experiments is to examine whether the desired properties (section \ref{sec:property}) hold. The order-invariant property is enforced by design. The size-flexibility property will be assessed by inspecting the sensitivity of the extraction and generative processes to the variability of the number of parts. The size-generality property will be assessed by inspecting the quality of the generative model when the number of parts increases significantly beyond the size range used during training. 

A last goal is to understand how CompVAE manages to store the information of the model in respectively the $w_i$s and $z$. The conjecture done (section \ref{sec:conjec}) was that the latent $w_i$s would take in charge the information of the parts, while the latent $z$ would model the interactions among the parts. The use of synthetic problems where the quantity of information required to encode the parts can be quantitatively assessed will permit to test this conjecture.
A related question is whether the generative model is able to capture the fact that {the whole is not the sum of its parts}. This question is investigated using non-linear perturbations, possibly operating at the whole and at the parts levels, and comparing the whole perturbed $x$ obtained from the $\ell_i$s, and the aggregation of the perturbed $x_i$s generated from the $\ell_i$ parts. The existence of a difference, if any, will establish the value of the CompVAE generative model compared to a simple Monte-Carlo simulator, independently sampling parts and thereafter aggregating them. 

\subsection{1D and 2D Proofs of concept}
Two synthetic problems have been considered to empirically answer the above questions.\footnote{These problems are publicly available at https://github.com/vberger/compvae .}

\def\C{{\mbox{$C$}}}
\paragraph{In the 1D synthetic problem,} the set $\Lambda$ of categories is a finite set of frequencies $\lambda_1 \ldots \lambda_{10}$. A given "part" (here, curve) 
is a sine wave defined by its frequency $\ell_i$ in $\Lambda$ and its intrinsic features, that is, its amplitude $a_i$ and phase $\kappa_i$.
The whole $x$ associated to $\{\ell_1, \ldots \ell_K \}$ is a finite sequence of size $T$, deterministically defined from the non-linear combination of the curves: 
$$x(t) = K \tanh\left( \frac{\C}{K} \sum_{i = 0}^K a_i \cos\left(\frac{2\pi \ell_i}{T} t + \kappa_i\right) \right)$$
with $K$ the number of sine waves in $x$, \C\  a parameter controlling the non-linearity of the aggregation of the curves in $x$, and $T$ a global parameter controlling the sampling frequency. For each part (sine wave), $a_i$ is sampled from $\mathcal{N}(1;0.3)$, and $\kappa_i$ is sampled from $\mathcal{N}\left(0; \frac{\pi}{2}\right)$. 

\begin{figure}
	\centering
	\resizebox{0.8\columnwidth}{!}{\input{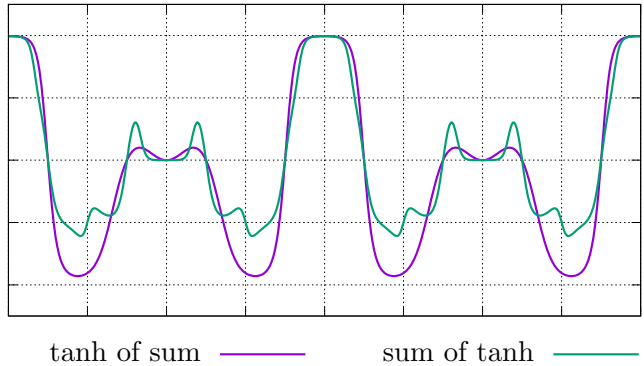}}
	\caption{Non-linear part-to-whole aggregation (purple) compared to the sum of non-linear perturbations of the parts (green). Better seen in color. Both curves involve a non-linear transform factor $\C = 3$.}
	\label{fig:non-linear-sines}
\end{figure}

The part-to-whole aggregation is illustrated on Fig. \ref{fig:non-linear-sines}, plotting the non-linear transformation of the sum of 4 sine waves, compared to the sum of non-linear transformations of the same sine waves.
The sensitivity to \C\ is illustrated in supplementary material (Appendix \ref{app:data-generation} Fig. \ref{fig:lambda-impact}).
\C\ is set to 3 in the experiments.

This 1D synthetic problem features several aspects relevant to the empirical assessment of CompVAE. 
Firstly, the impact of adding or removing one part can be visually assessed as it changes the whole curve:  the general magnitude of the whole curve is roughly proportional to its number of parts. Secondly, each part involves, besides its category $\ell_i$, some intrinsic variations of its amplitude and phase.  Lastly, the whole $x$ is not the sum of its parts (Fig. \ref{fig:non-linear-sines}).

The generative model $p_\theta(x | z , \sum_i w_i)$ is defined as a Gaussian distribution $\mathcal{N}(\mu; \Delta(\sigma))$, the vector parameters $\mu$ and $\sigma$ of which are produced by the neural network (architecture details in supplementary material, section \ref{app:sines-nn-structure}).

\begin{figure}
	\centering
	\includegraphics[width=0.8\columnwidth]{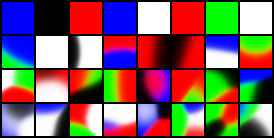}
	\caption{2D visual synthetic examples, including 1 to 4 sites (top to bottom). Note that when neighbor sites have same color, the image might appear to have been generated with less sites than it actually has.}
	\label{fig:color-gradients}
\end{figure}

\paragraph{In the 2D synthetic problem,} each category in $\Lambda$ is composed of one out of five colors ($\{red, green,$ $blue, white, black\}$) associated with a location $(x, y)$ in  $[0,1]\times[0,1]$. Each $\ell_i$ thus is a colored site, and its internal variability is its intensity. The whole $x$ associated to a set of $\ell_i$s is an image, where each pixel is colored depending on its distance to the sites and their intensity (Fig. \ref{fig:color-gradients}). 
Likewise, the observation model $p_\theta(x | z , \sum_i w_i)$ is a Gaussian distribution $\mathcal{N}(\mu; \Delta(\sigma))$, the parameters $\mu$ and $\sigma$ of which are produced by the neural network. The observation variance is shared for all three channel values (red, green, blue) of any given pixel.
Architecture details are given in supplementary material (section \ref{app:dots-nn-structure}).

The 2D problem shares with the 1D problem the fact that each part is defined from its category $\ell_i$ (resp. a frequency, or a color and location) on the one hand, and its specifics on the other hand (resp, its amplitude and frequency, or its intensity); additionally, the whole is made of a set of parts in interaction. However, the 2D problem is significantly more complex than the 1D, as will be discussed in section \ref{sec:morecomplex}.

\subsection{Experimental setting}
CompVAE is trained as a mainstream VAE, except for an additional factor of difficulty: the varying number of latent variables (reflecting the varying number of parts) results in a potentially large number of latent variables. This large size and the model noise in the early training phase can adversely affect the training procedure, and lead it to diverge. The training divergence is prevented using a batch size set to 256. 
The neural training hyperparameters are dynamically tuned using the Adam optimizer \cite{kingma_adam:_2014} with $\alpha=10^{-4}$, $\beta_1 = 0.5$ and $\beta_2 = 0.9$, which empirically provide a good compromise between training speed, network stability and good convergence. 
On the top of Adam, the annealing of the learning rate $\alpha$ is achieved, dividing its value by 2 every 20,000 iterations, until it reaches $10^{-6}$.

For both problems, the data is generated on the fly during the training,
\footnote{The data generator is given in supplementary material, section \ref{app:data-generation}.}
preventing the risk of overfitting. The overall number of iterations (batches) is up to\footnote{Experimentally, networks most often converge much earlier. } 500,000. The computational time on a GPU GTX1080 is 1 day for the 1D problem, and 2 days for the 2D problem. 

Empirically, the training is facilitated by gradually increasing the number $K$ of parts in the datapoints. Specifically, the number of parts is uniformly sampled in $[[1,K]]$ at each iteration, with $K = 2$ at the initialization and $K$ incremented by 1 every 3,000 iterations, up to 16 parts in the 1D problem and 8 in the 2D problem.

\section{CompVAE: Empirical Validation}

\label{sec:resu}

This section reports on the proposed proofs of concept of the CompVAE approach. 

\subsection{1D Proof of Concept}
\begin{figure}
	\centering
	\resizebox{\columnwidth}{!}{\input{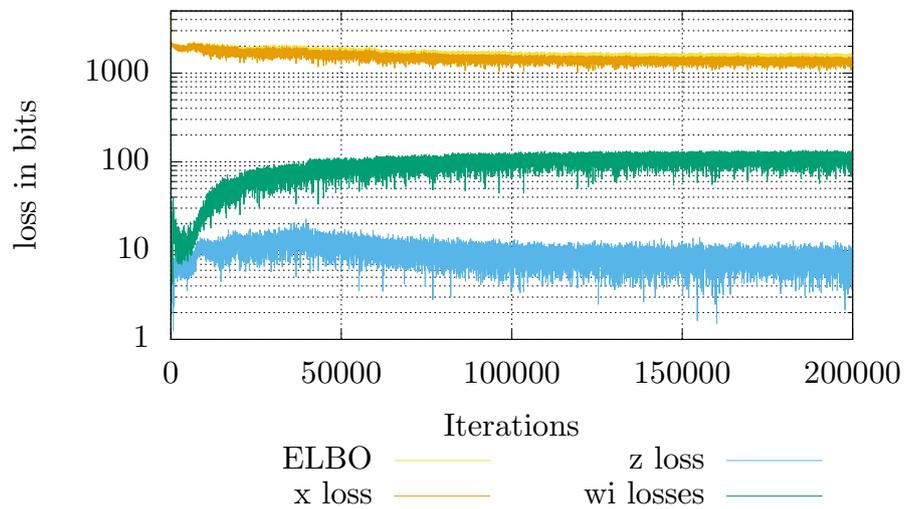}}
	\caption{CompVAE, 1D problem: Losses of the latent variables respectively associated to the parts ($w_i$, green), to the whole ($z$, blue), and the reconstruction loss of $x$ (yellow), in log scale. Better seen in color.}
	\label{fig:sines-train-curves}
\end{figure}

Fig. \ref{fig:sines-train-curves} displays in log-scale the losses of the $w_i$s and $z$ latent variables along time, together with the reconstruction loss and the overall ELBO loss summing the other three (Eq. (\ref{eq:elbo2})). The division of labor between the $w_i$s and the $z$ is seen as the quantity of information stored by the $w_i$s increases to reach a plateau at circa 100 bits, while the quantity of information stored by $z$ steadily decreases to around 10 bits. As conjectured (section \ref{sec:conjec}), $z$ carries little information.

Note that the $x$ reconstruction loss remains high, with a high ELBO even at convergence time, although the generated curves "look good". This fact is explained from the high entropy of the data: on the top of the specifics of each part (its amplitude and phase), $x$ is described as a $T$-length sequence: the temporal discretization of the signal increases the variance of $x$ and thus causes a high entropy, which is itself a lower bound for the ELBO.
Note that a large fraction of this entropy is accurately captured by CompVAE through the variance of the generative model $p_\theta(x | z, \wtil)$.

\begin{figure}
    \centering
    \resizebox{\textwidth}{!}{\input{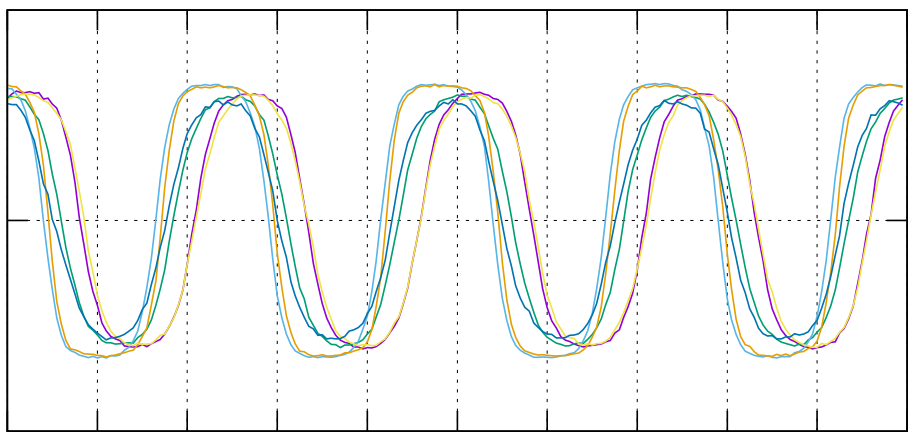}}
    \caption{1D Audio benchmark: Intrinsic variance of the parts (sine curves) generated by $p_\theta$ for a same value of $\ell_i$.}
    \label{fig:sines-variance}
\end{figure}

\begin{figure}
	\centering
	\resizebox{\columnwidth}{!}{\input{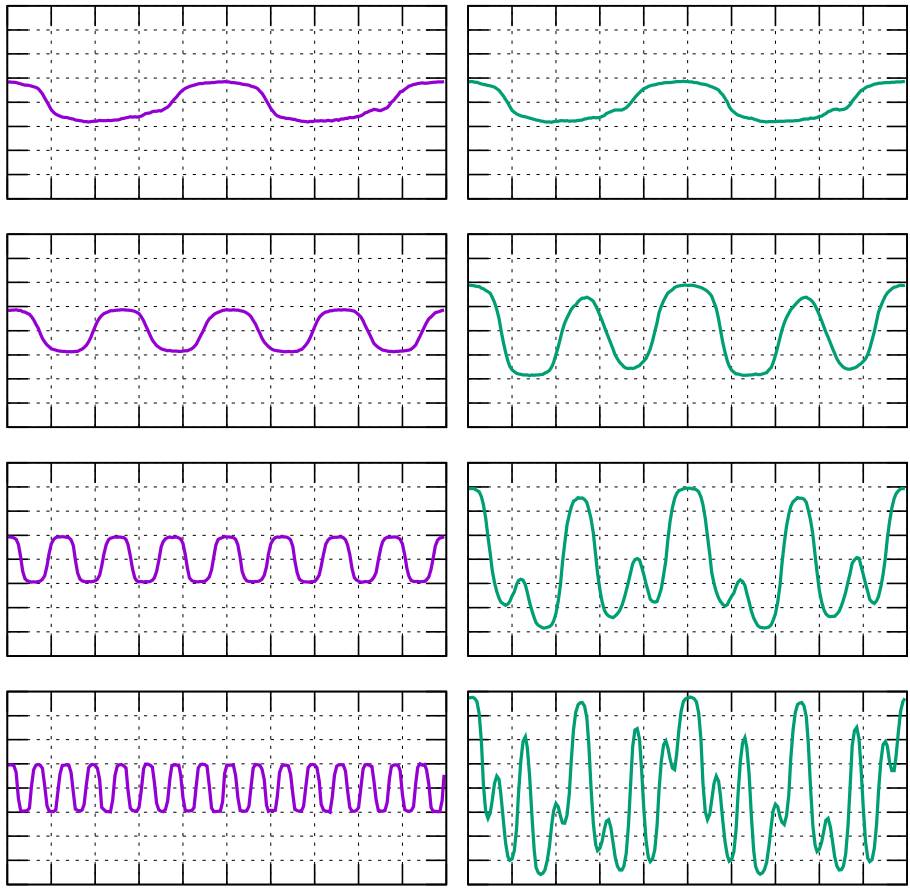}}
	\caption{CompVAE, 1D problem: Ensemblist recomposition of the whole (right column) from the parts (left column). On each row is given the part (left) and the whole (right) made of this part and all above parts. }
	\label{fig:sines-growing}
\end{figure}

The ability of "ensemblist disentanglement" is visually demonstrated on Fig. \ref{fig:sines-growing}: considering a set of $\ell_i$, the individual parts $w_i$ are generated (Fig. \ref{fig:sines-growing}, left) and gradually integrated to form a whole $x$  (Fig. \ref{fig:sines-growing}, right) in a coherent manner.

The size-generality property is satisfactorily assessed as the model could be effectively used with a number of parts $K$ ranging up to 30 (as opposed to 16  during the training) without requiring any re-training or other modification of the model (results omitted for brevity).

\subsection{2D Proof of Concept}\label{sec:morecomplex}
\begin{figure}
	\centering
	\resizebox{\columnwidth}{!}{\input{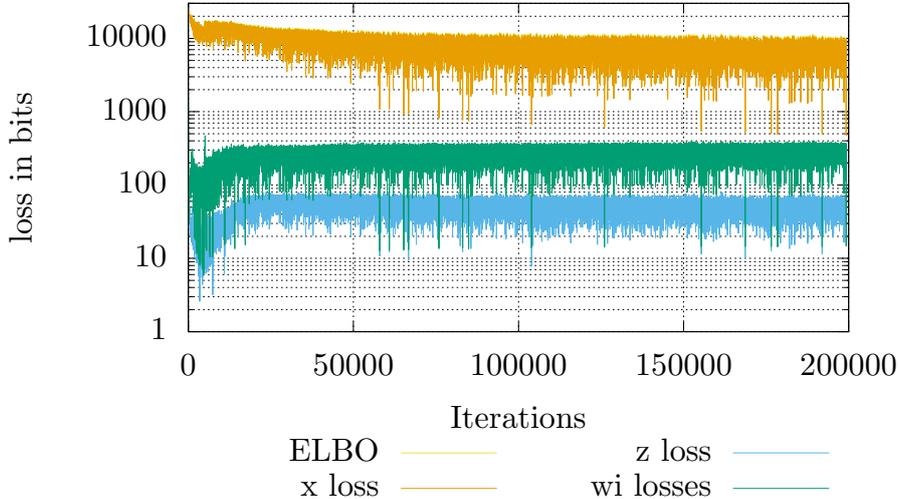}}
	\caption{CompVAE, 2D problem: Losses of the latent variables respectively associated to the parts ($w_i$, green), to the whole ($z$, blue), and the reconstruction loss of $x$ (yellow), in log scale. Better seen in color.}
	\label{fig:shades-train-curves}
\end{figure}

As shown in Fig. \ref{fig:shades-train-curves}, the 2D problem is more complex. On the one hand, a 2D part only has a local impact on $x$ (affecting a subset of pixels) while a 1D part has a global impact on the whole $x$ sequence. On the other hand, the number of parts has a global impact on the range of $x$ in the 1D problem, whereas each pixel value ranges in the same interval in the 2D problem. Finally and most importantly, $x$ is of dimension 200 in the 1D problem, compared to dimension $3,072$ ($3 \times 32 \times 32$) in the 2D problem. For these reasons, the latent variables here need to store more information, and the separation between the $w_i$ (converging toward circa 200-300 bits of information) and $z$ (circa 40-60 bits) is less clear.

Likewise,  $x$ reconstruction loss remains high, although the generated images "look good", due to the fact that the loss precisely captures the discrepancies in the pixel values that the eye does not perceive. 

\begin{figure}
	\centering
	\includegraphics[width=0.5\columnwidth]{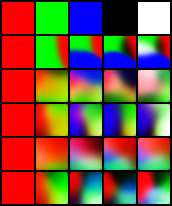}
	\caption{CompVAE, 2D problem. First row: parts $\ell_1 \ldots \ell_5$. Second row: the $i$-th square depicts the $x$ defined from $\ell_1$ to $\ell_i$ as generated by the ground truth. Rows 3-6: different realizations of the same combination by the trained CompVAE - see text. Best viewed in colors.}
	\label{fig:shades-growing}
\end{figure}

Finally, the ability of "ensemblist disentanglement" is inspected by incrementally generating the whole $x$ from a set of colored sites (Fig. \ref{fig:shades-growing}). The top row displays the colors of $\ell_1 \ldots \ell_5$ from left to right. On the second row, the $i$-th square shows an image composed from $\ell_1 \ldots \ell_i$ by the ground truth generator, and rows 3 to 6 show images generated by the model from the same $\ell_1 \ldots \ell_i$. While the generated $x$ generally reflects the associated set of parts, some advents of black and white glitches are also observed (for instance on the third column, rows 3 and 5). These glitches are blamed on the saturation of the network (as black and white respectively are represented as $(0,0,0)$ and $(1,1,1)$ in RGB), since non linear combinations of colors are used for a good visual rendering\footnote{Color blending in the data generation is done taking into account gamma-correction.}.

\section{Discussion and Perspectives}

\label{sec:discussion}
The main contribution of the paper is the generative framework CompVAE, to our best knowledge the first generative framework able to support the generation of data based on a multi-ensemble $\{ \ell_i \}$. Built on the top of the celebrated VAE, CompVAE\ learns to optimize the conditional distribution $p_\theta(x | \{\ell_i\})$ in a theoretically sound way, through introducing latent variables (one for each part $\ell_i$), enforcing their order-invariant aggregation and learning another latent variable to model the interaction of the parts. Two proofs of concepts for the approach, respectively concerning a 1D and a 2D problem, have been established with respectively very satisfactory and satisfactory results.

This work opens several perspectives for further research. A first direction in the domain of computer vision consists of combining CompVAE\ with more advanced image generation models such as PixelCNN \cite{van_den_oord_conditional_2016} in a way similar to PixelVAE \cite{gulrajani_pixelvae:_2016}, in order to generate realistic images involving a predefined set of elements along a consistent layout.

A second perspective is to make one step further toward the training of fully programmable generative models. The idea is to incorporate explicit biases on the top of the distribution learned from unbiased data, to be able to sample the desired sub-spaces of the data space. In the motivating application domain of electric consumption for instance, one would like to sample the global consumption curves associated with high consumption peaks, that is, to bias the generation process toward the top quantiles of the overall distribution.  

\section*{Acknowledgments}
This work was funded by the ADEME \#1782C0034 project {\em NEXT} (\url{https://www.ademe.fr/next}).

The authors would like to thank Balthazar Donon and Corentin Tallec for the many useful and inspiring discussions.

\bibliography{bibliography}

\clearpage
\appendix

\section{Model structures}

\subsection{NN structures for the 1d audio problem}
\label{app:sines-nn-structure}

The datapoints $x$ are 200-dimensional vectors each. We use a latent space with 128 dimensions for $z$, and a latent space with 256 dimensions for the $w_i$.

\subsubsection{Structure of the generator network $p_\theta$}

We model $p_\theta(w_i | l_i)$ as a learned embedding from a discrete value $l_i$ to the mean $\mu_{w_i}$ and log-variance $\nu_i$ of a distribution $\mathcal{N}(\mu_{w_i}; \sigma^2 = \exp{\nu_i})$.

The next layer, $p_\theta(z | \sum_i w_i)$ is modelled using a neural network taking $\sum_i w_i$ as input, and returning the mean $\mu_z$ and log-variance $\nu_z$ of a distribution $\mathcal{N}(\mu_z; \sigma^2 = \exp{\nu_z})$. Its structure is described in table \ref{tab:sines-struct-pz}.

\begin{figure}[h]
    \centering
    \input{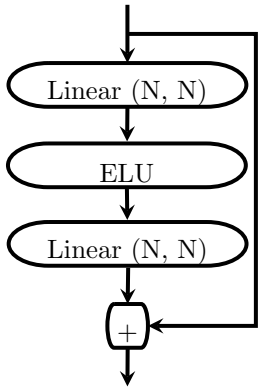}
    \caption{Definition of the residual blocks used in this paper.}
    \label{fig:residual-block}
\end{figure}

\begin{table}[h]
    \centering
    \begin{tabular}{r|l}
        \hline
        Layer & Activation \\
        \hline
        input: $\sum_i w_i$ & \\
        Linear(1024, 1280) & ELU \\
        Linear(1280, 512) & \\
        Reshape from $(512,)$ to $(2,256)$ & \\
        output $(\mu_{z}, \nu_{z})$ & \\
        \hline
    \end{tabular}
    \caption{Structure of the neural network modelling $p_\theta(z | \sum_i w_i)$.}
    \label{tab:sines-struct-pz}
\end{table}

The final layer, $p_\theta(x | z, \sum_i w_i)$ is modelled using a neural network taking $\sum_i w_i$ and $z$ as input, and returning a couple of vectors $(\mu_x, \nu_x)$, parametring a distribution $\mathcal{N}(\mu_x; \sigma^2 = \exp(\nu_x))$. The distribution is parametred by its log-variance rather than its standard deviation for stability reasons. The network structure is given as table \ref{tab:sines-struct-px}. Note that the transposed convolution layers final output is larger than the 200 neurons of the data and then cropped, this is to avoid boundary effects with transposed convolutions.

\begin{table}[h]
    \centering
    \begin{tabular}{r|l}
        \hline
        Layer & Activation \\
        \hline
        input: Concatenate($\sum_i w_i$, $z$) & \\
        Linear(1280, 800) & ELU \\
        Residual(800) & ELU \\
        Residual(800) & ELU \\
        Residual(800) & ELU \\
        Reshape from $(800,)$ to $(160,5)$ & \\
        TransposedConv1d(160, 80, ks=4, s=2, p=0) & ELU \\
        Conv1d(80, 80, ks=7, s=1, p=3) & ELU \\
        TransposedConv1d(80, 40, ks=8, s=4, p=0) & ELU \\
        Conv1d(40, 40, ks=7, s=1, p=3) & ELU \\
        TransposedConv1d(40, 20, ks=15, s=5, p=0) & ELU \\
        Conv1d(20, 2, ks=7, s=1, p=3) & \\
        Crop[:,35:235] & \\
        Output $(\mu_x, \nu_x)$ & \\
        \hline
    \end{tabular}
    \caption{Structure of the neural network modelling $p_\theta(x | z, \sum_i w_i)$. For the convolution layers, the parameters are: ks for kernel size, s for stride, and p for padding.}
    \label{tab:sines-struct-px}
\end{table}

\subsubsection{Structure of the inference network $q_\phi$}

The inference network features weight sharing between $q_\phi(z |x)$ and $q_\phi(w_i | ...)$: a preprocessing block for $x$, described in table \ref{tab:sines-preprocess-x}.

\begin{table}[h]
    \centering
    \begin{tabular}{r|l}
        \hline
        Layer & Activation \\
        \hline
        input: $x$ & \\
        Conv1d(1, 40, ks=10, s=5, p=0) & ELU \\
        Conv1d(40, 40, ks=7, s=1 p=3) & ELU \\
        Conv1d(40, 80, ks=6, s=3, p=0) & ELU \\
        Conv1d(80, 80, ks=7, s=1, p=3) & ELU \\
        Conv1d(80, 160, ks=4, s=2, p=0) & ELU \\
        Reshape from $(160,5)$ to $(800,)$ & \\
        Residual(800) & ELU \\
        \hline
    \end{tabular}
    \caption{Structure of the preprocessing block for $x$ in $q_\phi$. For the convolution layers, the parameters are: ks for kernel size, s for stride, and p for padding.}
    \label{tab:sines-preprocess-x}
\end{table}

The first half of the inference network, $q_\phi(z | x)$ is described in table \ref{tab:sines-struct-qz}, and the second half, $q_\phi(w_i | ...)$ in table \ref{tab:sines-struct-qw}. This second half is build using GraphBlocks, described in the main paper in section \ref{sec:q-model}, and whose neural network structure is described in table \ref{tab:graphblock}.

\begin{table}[h]
    \centering
    \begin{tabular}{r|l}
         \multicolumn{2}{c}{\bf model of $g$} \\
         \hline
         Layer & Activation\\
         \hline
         input: $h_j$ & \\
         Residual(N) & ELU \\
         Residual(N) \\
         \hline 
    \end{tabular}
    \begin{tabular}{r|l}
         \multicolumn{2}{c}{\bf model of $f$} \\
         \hline
         Layer & Activation\\
         \hline
         input: $\sum_{j\neq i} g(h_j)$ & tanh \\
         Residual(N) & \\
         Concatenate with $h_i$ & \\
         Linear(2N, M) & ELU \\
         Residual(M) & \\
         \hline 
    \end{tabular}
    \caption{Structure of a $GraphBlock(N \rightarrow M)$. Given $K$ feature vectors $h_i$ of
    size $N$, it computes $k$ output vectors of size $M$ $h'_i$ as so: $h'_i = f(h_i, \sum_{j \neq i} g(h_j))$.}
    \label{tab:graphblock}
\end{table}

\begin{table}[h]
    \centering
    \begin{tabular}{r|l}
        \hline
        Layer & Activation \\
        \hline
        input: pr($x$) & \\
        Linear(800, 512) & ELU \\
        Residual(512) & ELU \\
        Residual(512) & ELU \\
        Linear(512, 512) & \\
        Reshape from $(512,)$ to $(2,256)$ & \\
        output $(\mu_{z}, \nu_{z})$ & \\
        \hline
    \end{tabular}
    \caption{Structure of the network implementing $q_\phi(z | x)$. pr($x$) represents the output of the prepocessing block described previously}.
    \label{tab:sines-struct-qz}
\end{table}

\begin{table}[h]
    \centering
    \begin{tabular}{r|l}
        \hline
        Layer & Activation \\
        \hline
        input: (pr($x$), $z$, embedding($l_i$)) & \\
        GraphBlock(2080, 2048) & ELU \\
        GraphBlock(2048, 2048) & ELU \\
        GraphBlock(2048, 2048) & ELU \\
        Linear(2048, 3072) & ELU \\
        Reshape from $(3072,)$ to $(3, 1024)$ & \\
        output $(\mu_{w_i}, \nu_{w_i}, \rho_{w_i})$ & \\
        \hline
    \end{tabular}
    \caption{Structure of the network implementing $q_\phi(\{w_i\}|...)$. pr($x$) represents the output of the prepocessing block described previously. All layers apart from GraphBlocks are applied independently for each $w_i$ variable.}.
    \label{tab:sines-struct-qw}
\end{table}

\subsection{NN structures for the 2D color gradient problem}
\label{app:dots-nn-structure}

The datapoints are 32x32 RGB images each. We use latent space sizes of 2048 for the $w_i$, and $1024$ for $z$.

\subsubsection{Structure of the generator network $p_\theta$}

The first layer, $p_\theta(w_i | l_i)$ is modelled using a neural network taking $l_i$ as input and returning the mean $\mu_w{w_i}$ and log-variance $\nu_{w_i}$ of a distribution $\mathcal{N}(\mu_{w_i}; \sigma^2 = \exp{\nu_{w_i}})$. Its structure is described in table \ref{tab:blobs-struct-pw}.

\begin{table}[h]
    \centering
    \begin{tabular}{r|l}
        \hline
        Layer & Activation \\
        \hline
        input: $pos_i$ & \\
        Linear(2, 32) & ELU \\
        Concatenate with embedding or $col_i$ & \\
        Linear(64, 1024) & ELU \\
        Linear(1024, 4096) & \\
        Reshape from $(4096,)$ to $(2, 2048)$ & \\
        output $(\mu_{w_i}, \nu_{w_i})$ & \\
        \hline
    \end{tabular}
    \caption{Structure of the network implementing $p_\theta(w_i|l_i)$. We split $l_i$ into its discrete part describing the color $col_i$ and its continuous part describing the location $loc_i$.}.
    \label{tab:blobs-struct-pw}
\end{table}

Then, $p_\theta(z | \sum_i w_i)$ is modelled using a neural network taking $\sum_i w_i$ as input and returning the mean $\mu_z$ and log-variance $\nu_z$ of a distribution $\mathcal{N}(\mu_z; \sigma^2 = \exp{\nu_z})$. Its structure is described in table \ref{tab:blobs-struct-pz}.

\begin{table}[h]
    \centering
    \begin{tabular}{r|l}
        \hline
        Layer & Activation \\
        \hline
        input: $\sum_i wi$ & \\
        Linear(2048, 1024) & ELU \\
        Residual(1024) & ELU \\
        Linear(1024, 2048) & \\
        Reshape from $(2048,)$ to $(2, 1024)$ & \\
        output $(\mu_{z}, \nu_{z})$ & \\
        \hline
    \end{tabular}
    \caption{Structure of the network implementing $p_\theta(z | \sum_i w_i)$.}
    \label{tab:blobs-struct-pz}
\end{table}

Finally, $p_\theta(x | z, sum_i w_i)$ is modelled using a neural network taking $z$ and $\sum_i w_i$ as input and return the mean $\mu_x$ and log variance $\nu_x$ of a distribution $\mathcal{N}(\mu_x, \sigma^2 = \exp(\nu_x))$. Its structure is described in table \ref{tab:blobs-struct-px}.

\begin{table}[h]
    \centering
    \begin{tabular}{r|l}
        \hline
        Layer & Activation \\
        \hline
        input: $\sum_i w_i$, $z$ & \\
        Residual(2048) on $\sum_i w_i$ & \\
        Linear(1024, 2048) on $z$ & \\
        Sum the two previous results & ELU \\
        Residual(2048) & tanh \\
        Residual(2048) & ELU \\
        Reshape from $(2048,)$ to $(128,4,4)$ & \\
        Bilinear upscaling x2 & \\
        Conv2d(128, 64, ks=5, s=1, p=2) & ELU \\
        Bilinear upscaling x2 & \\
        Conv2d(64, 32, ks=5, s=1, p=2) & ELU \\
        Bilinear upscaling x2 & \\
        Conv2d(32, 24, ks=5, s=1, p=2) & ELU \\
        Conv2d(16, 4, ks=5, s=1, p=2) & ELU \\
        Split into $(3,64,64)$ and $(1,64,64)$ & \\
        output $(\mu_{x}, \nu_x)$ & \\
        \hline
    \end{tabular}
    \caption{Structure of the network implementing $p_\theta(z | \sum_i w_i)$. For the convolution layers, the parameters are: ks for kernel size, s for stride, and p for padding.}
    \label{tab:blobs-struct-px}
\end{table}

\subsubsection{Structure of the inference network $q_\phi$}

The inference network features weight sharing between $q_\phi(z |x)$ and $q_\phi(w_i | ...)$: a preprocessing block for $x$, described in table \ref{tab:sines-preprocess-x}.

\begin{table}[h]
    \centering
    \begin{tabular}{r|l}
        \hline
        Layer & Activation \\
        \hline
        input: $x$ & \\
        Conv2d(3, 16, ks=5, s=1, p=2) & ELU \\
        Conv2d(16, 32, ks=4, s=2 p=1) & ELU \\
        Conv2d(32, 32, ks=5, s=1, p=2) & ELU \\
        Conv2d(32, 48, ks=4, s=2, p=1) & ELU \\
        Conv2d(48, 64, ks=4, s=2, p=1) & ELU \\
        Reshape from $(64,4,4)$ to $(1024,)$ & \\
        Residual(1024) & ELU \\
        \hline
    \end{tabular}
    \caption{Structure of the preprocessing block for $x$ in $q_\phi$. For the convolution layers, the parameters are: ks for kernel size, s for stride, and p for padding.}
    \label{tab:blobs-preprocess-x}
\end{table}

The first part of the inference network, $q_\phi(z|x)$ is described in table \ref{tab:blobs-struct-qz}, and the next layer, $q_\phi(\{w_i\} | x, z, \{\ell_i\})$ in table \ref{tab:blobs-struct-qw}.

\begin{table}[h]
    \centering
    \begin{tabular}{r|l}
        \hline
        Layer & Activation \\
        \hline
        input: pr($x$) & \\
        Residual(1024) & ELU \\
        Linear(1024, 2048) & \\
        Reshape from $(2048,)$ to $(2, 1024)$ & \\
        output $(\mu_{z}, \nu_{z})$ & \\
        \hline
    \end{tabular}
    \caption{Structure of the network implementing $q_\phi(z | x)$. pr($x$) is the output of the preprocessing layer described previously.}
    \label{tab:blobs-struct-qz}
\end{table}

\begin{table}[h]
    \centering
    \begin{tabular}{r|l}
        \hline
        Layer & Activation \\
        \hline
        input: $pos_i$ & \\
        Linear(2, 32) & ELU \\
        Concatenate with embedding of $col_i$, $z$ and pre($x$) & \\
        GraphBlock(2112, 2048) & ELU \\
        GraphBlock(2048, 2048) & ELU \\
        GraphBlock(2048, 2048) & ELU \\
        Linear(2048, 6144) & \\
        Reshape from $(6144,)$ to $(3, 2048)$ & \\
        output $(\mu_{w_i}, \nu_{w_i}, \rho_{w_i})$ & \\
        \hline
    \end{tabular}
    \caption{Structure of the network implementing $q_\phi(\{w_i\} | x, z, \{\ell_i\})$. }
    \label{tab:blobs-struct-qw}
\end{table}

\clearpage

\section{Data generation}

\label{app:data-generation}

\subsection{Data generation for the 1d audio problem}

\begin{lstlisting}[language=Python,basicstyle=\tiny]
import torch
import math
import random

def generate_curves(freqs, timesteps, resolution, C):
    """
    Generate a batch of curves with specified characteristics:
    - freqs: chosen frequencies as an int array of dimensions
      [batchlen, number of sines]
    - timesteps: total number of sampling points per example
    - resolution: number of sampling points per fundamental
      period
    - C: non-linearity factor of the combination
    
    Returns a torch array of size [batchlen, timesteps] with
    the data
    """
    (batchlen, nfreqs) = freqs.size()
    freqs = freqs.view(batchlen, 1, nfreqs).float()
    amplitudes = freqs.new_empty(freqs.size()).normal_(1.0, 0.3)
    phases = freqs.new_empty(freqs.size()).normal_(0.0, 0.8)
    times = torch.arange(0.0, timesteps, device=freqs.device)
    times /= resolution
    times = times.view(1, timesteps, 1)
    curve = torch.sum(
        amplitudes * torch.cos(2*math.pi*freqs*time + phases),
        dim=2
    )
    return nfreqs * torch.tanh(C * curve / nfreqs)

def generate_batch(batchlen, freqrg=(1,10), nfreqrg=(1,16),
                   timesteps=200, resolution=100, device=None):
    """
    Generate a random batch according to the specified
    characteristics:
    - batchlen: number of examples in the batch
    - freqrg: the inclusive range of possible frequencies
    - nfreqrg: the inclusive range of possible number of
      sines per example
    - timesteps: total number of sampling points per example
    - resolution: number of sampling points per fundamental
      period
    - device: the generation can be made directly on the GPU
      for increased speed
    
    Returns a tuple of:
    - a torch integer array of size [batchlen,n] containing
      the list of frequencies present in each example
    - a torch array of size [batchlen, timesteps] with the
      data
    """
    nfreq = random.randint(nfreqrng[0], nfreqrng[1])
    freqs = torch.randint(freqrng[0], freqrng[1],
        size=(batchlen, nfreq), device=device)
    return (freqs,
        generate_curves(freqs, timesteps, resolution))
\end{lstlisting}

\begin{figure}
    \centering
    \resizebox{\columnwidth}{!}{\input{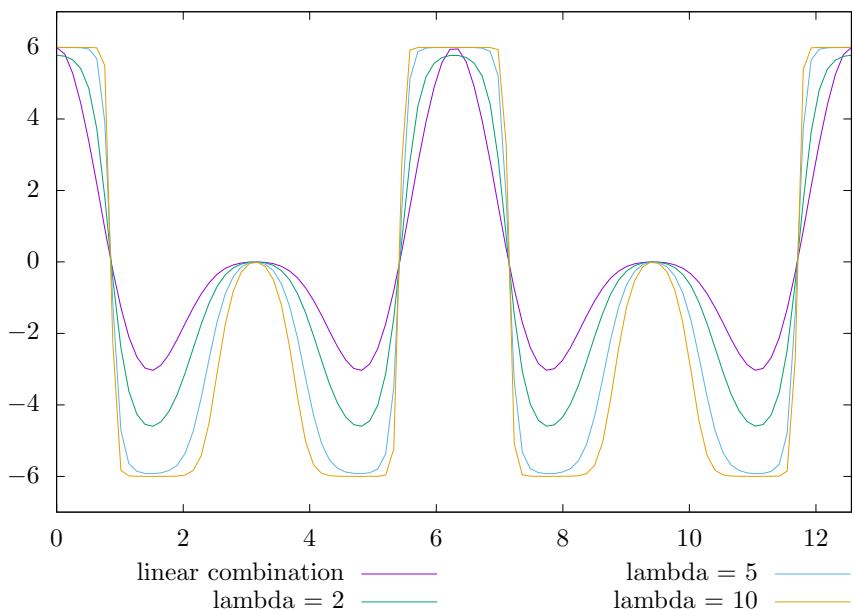}}
    \caption{Impact of the \C\ factor on the part-to-whole aggregation of the sine curves, compared to a linear aggregation.}
    \label{fig:lambda-impact}
\end{figure}

Figure \ref{fig:lambda-impact} illustrates the impact of varying the \C\ parameter in the generation of the data for the 1D problem.

\subsection{DATA GENERATION FOR THE DOTS PROBLEM}

\begin{lstlisting}[language=Python,basicstyle=\tiny]
import torch
import math
import random

COLORS = torch.tensor([
        [1.0, 0.0, 0.0], # RED
        [0.0, 1.0, 0.0], # GREEN
        [0.0, 0.0, 1.0], # BLUE
        [0.0, 0.0, 0.0], # BLACK
        [1.0, 1.0, 1.0], # WHITE
])

def draw_gradient(locations, colors):
    """
    Generate a color gradient from given anchor points:
    - locations: a [batchlen, num_points, 2] sized array
      containing the coordinates of each anchor point ranging
      in [-1, 1]
    - colors: a [batchlen, num_points, 3] sized array
      containing the RGB color associated with each anchor point
     
    Returns a torch array of size [batchlen, 3, 32, 32]
    containing the images.
    """
    batchlen = locations.size(0)
    K = locations.size(1)
    xs = locations[:,:,0:1]
    ys = locations[:,:,1:2]
    # make a range from -1 to 1 matching the pixels
    line = torch.arange(32, device=locations.device).float()
    line = line * 2.0 / 31.0 - 1.0
    line = line.view(1,1,32)
    x_distance = ((line-xs)**2).view(batchlen,K,1,32)
    y_distance = ((line-ys)**2).view(batchlen,K,32,1)
    distance_grid = x_distance + y_distance
    # the intensity factor measures how quickly the color from
    # an anchor point is replaced by its neighbors
    anchor_intensity = distance_grid
                        .new_empty((batchlen, K, 1, 1))
                        .uniform_(5, 10)
    intensity = torch.softmax(- anchor_intensity * distance_grid,
        dim=1)
    intensity = intensity.intensity.view(batchlen, K, 1, 32, 32)
    colors = colors.view(batchlen, K, 3, 1, 1)
    return torch.sum(intensity * colors ,dim=1)

def generate_batch(batchlen, anchorcount, device=None):
    """
    Generate a batch according to the specified characteristics:
    - batchlen: the size of the batch
    - anchorcount: the number of anchor points in each image
    - device: the generation can be made directly on the GPU
      for increased speed
    
    Returns a tuple of:
    - a torch array of dimensions [batchlen, 3, 32, 32]
      containing the images
    - an integer torch array of dimensions
      [batchlen, anchorcount] containing the color of each
      anchor point
    - a torch array of dimensions [batchlen, anchorcount, 2]
      containing the coordinates of each anchor point in the
      images
    """
    (amin, amax) = anchorcount
    num_anchors = random.randrange(amin, amax+1)
    img = torch.ones((batchlen, 3, 32, 32), device=device)
    labels = torch.randint(COLORS.shape[0],
        size=(batchlen, num_anchors), device=device)
    locations = torch.rand(batchlen, num_blobs, 2, device=device)
    # ranging from -1.9 to +1.9 so that anchor points are
    # never on the image border
    locations = locations * 1.8 - 0.9
    img = draw_gradient(locations,
        COLORS.to(device=device)[labels, :])
    # Gamma-correct the image to create nice color gradients
    return (torch.clamp(img ** (1/2.4), 0.0, 1.0),
        labels, locations)


\end{lstlisting}

\clearpage
\section{Multivariate Gaussian parametrization}

\label{app:multivariate-parameter}

\subsection{Parametrization definition}

In order to define a join distribution for the $w_i$ variables, we will work by correlating them dimension wise. Here, $w_{i,j}$ represents the $j$-th coordinate of $w_i$.

For a given coordinate $j$, we model $(w_{1,j}, w_{2,j}, .. w_{K,j}$ as a $K$-dimensional multivariate normal distribution defined by three vectors: $\mu_{i,j}$, $\sigma_{i,j} > 0$ and $0 < \rho_{i,j} < 1$ and the following sampling process. A vector $\epsilon_{i,j}$ is sampled from $\mathcal{N}(0;1)$, and $w_{i,j}$ is computed as:

\begin{equation}
    w_{i,j} = \mu_{i,j} + \sigma_{i,j}\left(\epsilon_{i,j} - \rho_{i,j} \sum_{i'=1}^K \epsilon_{i',j}\right)
\end{equation}

Expressed in matrix form, if we set $D_j = Diag(\sigma_{1,j}, .. , \sigma_{K,j})$ and $S_j = I - \rho_j \mathbf{1}^T$ where $\rho_j$ is the column vector $(\rho_{1,j}, .. , \rho_{K,j})$ and $\mathbf{1}^T$ is the line vector $(1, .., 1)$, the vector $(w_{1,j}, .. , w_{K,j})$ is sampled from the normal distribution of mean $(\mu_{1,j}, .. , \mu_{K,j})$ and of covariance matrix $D_j S_j S_j^T D_j^T$.

The motivation of such a parametrization is based on the fact that the inference network needs to control $\Sigma_i w_i$ in order to ensure a good reconstruction by the VAE. Using this parametrization, we can see that:

\begin{equation}
    \mathrm{Var}\left( \sum_{i=1}^{K} w_{i,j} \right) = \left(\sum_{i=1}^{K} \sigma_i^2\right)\left(1 - \sum_{i=1}^{K}\rho_{i,j}\right)
\end{equation}

The network can bring the variance of the sum of the $w_{i,j}$ arbitrarily close to $0$ by bringing the sum of the $\rho_{i,j}$ close to 1. To ensure the density $q_\phi(\{w_i\}|x,z,\{\ell_i\})$ remains well-defined, we must keep their sum strictly smaller than 1, which we achieve by using a softmax-like parametrization. Let use denote $\tilde{\rho}_{i,j}$ the pre-activation value associated to $\rho_{i,j}$, then:

\begin{equation}
    \rho_{i,j} = \frac{\exp(\tilde{\rho}_{i,j})}{1 + \sum_{i'=1}^{K} \exp(\tilde{\rho}_{i',j})}
\end{equation}

\subsection{Loss computation}

This parametrization also allows closed-form analytically computation of the Kullback-Leibler divergence between $q_\phi(\{w_i\}|x, z, \{\ell_i\})$ and $\prod_i p_\theta(w_i|\ell_i)$.

Indeed, one can exactly compute that $|D_j| = \prod_{i=1}^{K} \sigma_{i,j}$ and, using the determinant lemma, that $|S_j| = 1 - \sum_{i=1}^K \rho_{i,j}$.

Furthermore, given that the distribution associated with $p_\theta(w_{1,j}, .. w_{K,j})$ is a diagonal Gaussian, one can exactly compute relevant part of the loss, which is the Kullback-Leibler divergence between the two distributions:

\begin{equation}
    \begin{split}
        \mathcal{L}_{w} & = \frac{1}{2}\sum_{i=1}^K \left( \log\frac{\sigma_{p,i,j}^2}{\sigma_{q,i,j}^2} + \frac{(\mu_{p,i,j} - \mu_{q,i,j})^2}{\sigma_{p,i,j}^2}\right) \\
        & + \sum_{i=1}^K (1 - 2\rho_{i,j} + K \rho_{i,j}^2) \frac{\sigma_{q,i,j}^2}{\sigma_{p,i,j}^2} \\
        & - \log\left(1 - \sum_{i=1}^K \rho_{i,j}\right) \\
    \end{split}
\end{equation}

\end{document}